\pdfoutput=1
\documentclass[11pt]{article}

\usepackage[preprint]{acl}
\usepackage{hyperref}
\usepackage{times}
\usepackage{latexsym}

\usepackage[T1]{fontenc}
\usepackage[utf8]{inputenc}
\usepackage{amsmath}

\usepackage{microtype}

\usepackage{inconsolata}

\usepackage{graphicx}

\usepackage{multirow}
\usepackage{booktabs}
\usepackage{graphicx}

\title{Contradiction to Consensus: Dual‑Perspective, Multi‑Source Retrieval-Based Claim Verification with Source‑Level Disagreement using LLM}

\author{
  Md Badsha Biswas \\
  George Mason University \\
  Fairfax, VA, USA \\
  \texttt{mbiswas2@gmu.edu}
  \And
  Özlem Uzuner \\
  George Mason University \\
  Fairfax, VA, USA \\
  \texttt{ouzuner@gmu.edu}
}

\begin{document}
\maketitle
\begin{abstract}
The spread of misinformation across digital platforms can pose significant societal risks. Claim verification, a.k.a., fact-checking, systems can help identify potential misinformation. However, their efficacy is limited by the knowledge sources that they rely on. Most automated claim verification systems depend on a single knowledge source and utilize the supporting evidence from that source; they ignore the disagreement of their source with others. This limits their knowledge coverage and transparency. To address these limitations, we present a novel system for open-domain claim verification (ODCV) that leverages large language models (LLMs), multi-perspective evidence retrieval, and cross-source disagreement analysis. Our approach introduces a novel retrieval strategy that collects evidence for both the original and the negated forms of a claim, enabling the system to capture supporting and contradicting information from diverse sources: Wikipedia, PubMed, and Google. These evidence sets are filtered, deduplicated, and aggregated across sources to form a unified and enriched knowledge base that better reflects the complexity of real-world information. This aggregated evidence is then used for claim verification using LLMs. We further enhance interpretability by analyzing model confidence scores to quantify and visualize inter-source disagreement. Through extensive evaluation on four benchmark datasets with five LLMs, we show that knowledge aggregation not only improves claim verification but also reveals differences in source-specific reasoning. Our findings underscore the importance of embracing diversity, contradiction, and aggregation in evidence for building reliable and transparent claim verification systems. Our full code is available on GitHub \footnote{\href{https://anonymous.4open.science/r/Automated-Fact-Verification-system--0BF7/README.md}{https://anonymous.4open.science/r/Automated-Fact-Verification-system--0BF7/}}
\end{abstract}

\begin{center}
   \begin{figure*}
        \centering
        \includegraphics[width=\linewidth]{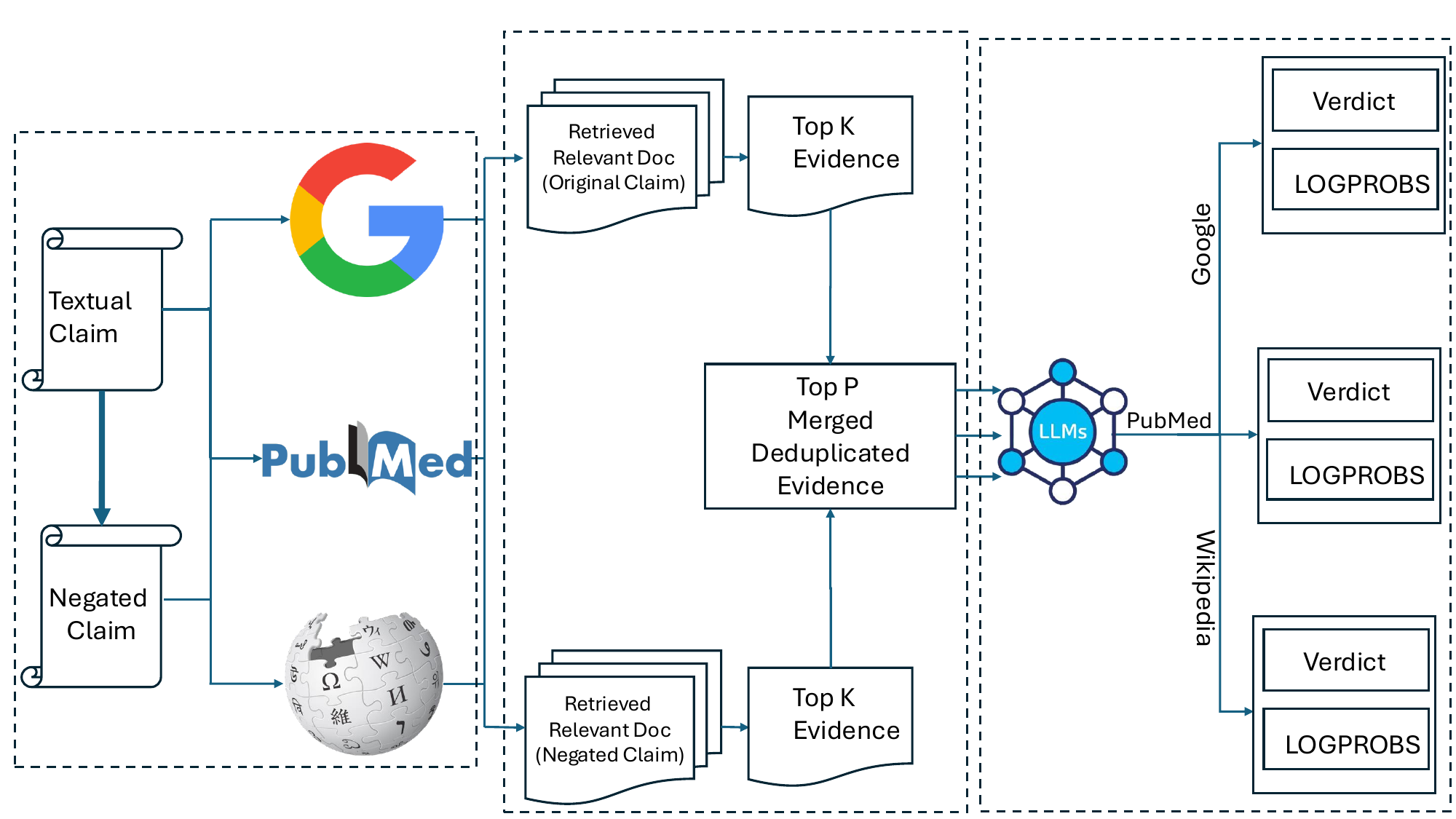} % Adjust the width
        \caption{Workflow of the proposed system. A claim and its negation are used to retrieve sentence-level evidence from Wikipedia, PubMed, and Google; candidate sentences are ranked and deduplicated, keeping the top-$p$ per source. The resulting candidate sentences from each source are aggregated into a cumulative evidence set $E_i$, which a zero-shot LLM uses to verify claims; log-probabilities for each source are reported and visualized to convey agreement and uncertainty.}
        \label{fig:proposed_system}
    \end{figure*}
\end{center}

\section{Introduction}

In an age where information travels faster than ever \cite{vosoughi2018spread}, misinformation \cite{barve2023multi} and disinformation \cite{ghosal2020resco} have emerged as pressing challenges for society. With just a few clicks, misleading claims and fabricated narratives can cascade across digital platforms, shaping public opinion and, in some cases, threatening lives and livelihoods \cite{arcos2022responses}. Nowhere is this danger more pronounced than in domains such as healthcare, finance, and public safety where the consequences can be dire \cite{sarrouti-etal-2021-evidence-based, rangapur2025fin, addy2020art}.

Natural language processing (NLP) can assist in automatic claim verification, which aims to detect misinformation and disinformation. Over the past few years, researchers have proposed increasingly advanced claim verification methods \cite{augenstein2024factuality,eldifrawi2024automated,wolfe2024impact}. However, most current systems focus narrowly on isolated pieces of evidence, overlooking the complex reality that information is often distributed across multiple sources rather than contained within a single repository. 

Moreover, the vast landscape of available knowledge is rarely leveraged to its full potential. Claim verification systems tend to rely on a single, primary knowledge source such as Wikipedia while ignoring the wealth of information contained in secondary or tertiary sources, like scientific literature or web search results. This limitation not only constrains the system’s ability to make well-informed decisions but also hinders interpretability for end-users, who are left unaware of the underlying disagreements among sources.

These challenges raise two pivotal questions at the heart of our work: How can we build claim verification systems that embrace, rather than ignore, the diversity and disagreement inherent in real-world evidence? And how can we transparently convey this complexity to users, empowering them to make informed decisions based on a full spectrum of perspectives? This study addresses these questions directly by introducing a novel system dedicated to open-domain claim verification. It effectively captures, measures, and conveys the complexity and contentious nature of knowledge in the digital era. Our methodology excels in retrieving context-specific evidence while also maintaining a diversity of viewpoints by including both supporting and opposing information. Our main contributions include:

\begin{itemize}
  
  \item We introduce a novel dual-perspective, multi-source, retrieval-based claim verification approach that considers both supporting and opposing evidence (i.e., dual perspective) of a claim from multiple sources. We evaluate our approach on four benchmark datasets, using evidence from three major knowledge bases.

  \item We demonstrate the effectiveness of aggregating evidence from multiple knowledge sources, as opposed to relying on a single source, for claim verification.

  \item Finally, we compare results across knowledge sources. We quantify the level of disagreement among the sources, even when the sources reach the same decision, using confidence scores.

\end{itemize}

We evaluate five state-of-the-art zero-shot large language models on claim verification. We achieve strong verification performance that is coupled with disagreements among sources, improving interpretability.

\section{Related Work}
Open-domain Claim verification (ODCV), a.k.a., Open-domain fact verification (ODFV), has grown ~\cite{patwa2022benchmarking,guo2022survey} through significant advances in evidence retrieval, model reasoning, and integration of diverse knowledge sources~\cite{dmonte2024claim}. While remarkable progress has been made, important challenges remain regarding evidence selection, handling contradictory evidence, and ensuring interpretability.

\subsection{Dual-Perspective Evidence Retrieval}
Most ODCV approaches focus only on evidence supporting a given claim, risking confirmation bias~\cite{zhou2019gear,hanselowski2018ukp,wang2017evidence,jiang2020hover}. CONFLICTINGQA shows that retrieval-augmented LMs often prioritize relevance over stylistic or credibility cues, diverging from human judgments~\cite{wan2024evidence}. Human-computer interaction and argumentation research study how people handle conflicting information~\cite{fogg2003users,kakol2017understanding,gretz2020large,toledo2019automatic}; however, reconciliation of contradictions for automated claim verification remains underexplored~\cite{wan2024evidence}. Thorne et al.~\cite{thorne2018automated} advocate considering both supporting and refuting evidence, which improves transparency but complicates reconciliation and verification result aggregation. Samarinas et al.~\cite{samarinas2021improving} increase explainability by distinguishing support vs.\ refute, yet retrieve evidence only using the original claim. We extend this paradigm by using both the original claim and its explicit negation for retrieval, forming a dual‑perspective evidence pool that better captures support \textit{and} contradiction for claim verification.

\subsection{Multi-Source Evidence Aggregation}

The choice of knowledge source and retrieval method strongly affects ODCV performance. ~\cite{vladika2024comparing} shows that both source (PubMed, Wikipedia, Google) and retrieval technique (BM25 vs.\ semantic search) significantly impact accuracy: PubMed is strongest for specialized biomedical claims, while Wikipedia better serves everyday health queries. Other work largely adopts single-source solutions: ~\cite{santos2020fact} uses a Portuguese Wikipedia-based knowledge graph and Google snippets; many systems rely on Wikipedia and cast verification as NLI~\cite{nie2019combining,si2021topic,thorne2018fact,yoneda2018ucl,ma2019sentence}; DrQA uses Wikipedia exclusively for open-domain QA~\cite{chen2017reading}; MultiFC retrieves web evidence via a search API~\cite{augenstein2019multifc}; and Cao et al.~\cite{cao2024multi} incorporate external multimodal signals with a heterogeneous graph. Despite this progress, there has been little to no systematic study of aggregating evidence across multiple independent sources for a single claim. We address this gap with a multi-source solution that retrieves from Wikipedia, PubMed, and Google, then deduplicates, merges, and ranks sentences to form a unified evidence set per claim, capturing both supporting and contradicting evidence that any single source may miss.

\subsection{Quantifying Disagreement Across Knowledge Sources}

Disagreement across sources is both common and informative in ODFV. Prior work quantifies uncertainty via inter‑source (or annotator) agreement~\cite{kavtaradze2024dominant}, models' ambiguity and annotator disagreement with soft labels (AMBIFC)~\cite{glockner2024ambifc}, promotes representational diversity via disagreement regularization in attention~\cite{li2018multi}, and argues for preserving divergent judgments rather than collapsing to majority vote~\cite{leonardelli2023semeval}. Token‑level uncertainty methods such as CCP further isolate uncertainty tied to factual content~\cite{fadeeva2024fact}. Yet, most systems still do not explicitly expose source‑level disagreement to users. In our approach, for each claim, we compute per‑source confidence scores (log‑probabilities) for the predicted verification label, then quantify dispersion across sources. Low dissemination indicates agreement; high dissemination flags disagreement and potential uncertainty. We visualize these per‑source logprobs to show confident agreements and ambiguous cases, enabling transparent, multi‑source claim verification interpretation.

\begin{table*}[]
\centering
\begin{tabular}{@{}cccccccccccc@{}}
\toprule
\multirow{2}{*}{Dataset}                 & \multirow{2}{*}{Model}                                                                & \multirow{2}{*}{\begin{tabular}[c]{@{}c@{}}Knowledge\\ Source\end{tabular}} & \multicolumn{4}{c}{Original Claim} &  & \multicolumn{4}{c}{Original + Negated Claim}                            \\ \cmidrule(lr){4-7} \cmidrule(l){9-12}  
                                         &                                                                                       &                                                                             & A       & P      & R      & F1     &  & A               & P               & R                 & F1              \\ \midrule
\multirow{15}{*}{\rotatebox{90}{SCIFACT}}
    & \multirow{3}{*}{\rotatebox{90}{\begin{tabular}{c}Llama\\ 70B\end{tabular}}}
        & Wikipedia                                                                   & 0.430   & 0.468  & 0.407  & 0.415  &  & 0.230 & 0.376           & 0.335 & 0.203           \\
    &                                                                             & Pubmed                                                                      & 0.597   & 0.588  & 0.597  & 0.584  &  & 0.617 & 0.609 & 0.626  & 0.605 \\
    &                                                                             & Google                                                                      & 0.550   & 0.543  & 0.558  & 0.530  &  & 0.607           & 0.615           & 0.620             & 0.573           \\ \cmidrule(l){2-12} 
    & \multirow{3}{*}{\rotatebox{90}{\begin{tabular}{c}Llama\\ 405B\end{tabular}}}
        & Wikipedia                                                                   & 0.447   & 0.489  & 0.420  & 0.425  &  & 0.443           & 0.475           & 0.421             & 0.427           \\
    &                                                                             & Pubmed                                                                      & 0.593   & 0.583  & 0.592  & 0.576  &  & 0.617           & 0.606           & 0.622             & 0.599           \\
    &                                                                             & Google                                                                      & 0.580   & 0.573  & 0.577  & 0.550  &  & 0.597           & 0.591           & 0.602             & 0.562           \\ \cmidrule(l){2-12} 
 & \multirow{3}{*}{\rotatebox{90}{Phi-4}}                                                & Wikipedia                                                                   & 0.410   & 0.474  & 0.391  & 0.379  &  & 0.413           & 0.473           & 0.401             & 0.400           \\
                                          &                                                                                       & Pubmed                                                                      & 0.583   & 0.574  & 0.583  & 0.578  &  & 0.587           & 0.573           & 0.599             & 0.579           \\
                                          &                                                                                       & Google                                                                      & 0.590   & 0.579  & 0.583  & 0.574  &  & 0.593           & 0.584           & 0.608             & 0.568           \\ \cmidrule(l){2-12} 
                                          & \multirow{3}{*}{\rotatebox{90}{Qwen 2.5}}                                             & Wikipedia                                                                   & 0.437   & 0.504  & 0.410  & 0.401  &  & 0.423           & 0.478           & 0.403             & 0.398           \\
                                          &                                                                                       & Pubmed                                                                      & 0.583   & 0.573  & 0.575  & 0.573  &  & 0.597           & 0.588           & 0.603             & 0.591           \\
                                          &                                                                                       & Google                                                                      & 0.587   & 0.578  & 0.577  & 0.565  &  & 0.590           & 0.567           & 0.595             & 0.563           \\ \cmidrule(l){2-12} 
                                          & \multirow{3}{*}{\rotatebox{90}{Mistral}}                                              & Wikipedia                                                                   & 0.393   & 0.447  & 0.368  & 0.358  &  & 0.393           & 0.438           & 0.375             & 0.368           \\
                                          &                                                                                       & Pubmed                                                                      & 0.573   & 0.565  & 0.571  & 0.566  &  & 0.590           & 0.585           & 0.595             & 0.585           \\
                                          &                                                                                       & Google                                                                      & 0.583   & 0.576  & 0.582  & 0.568  &  & 0.603           & 0.591           & 0.620             & 0.586           \\ 

 \midrule
\multirow{15}{*}{\rotatebox{90}{Averitec}}
    & \multirow{3}{*}{\rotatebox{90}{\begin{tabular}{c}Llama\\ 70B\end{tabular}}}
        & Wikipedia                                                                   & 0.259                     & 0.417                     & 0.355                     & 0.229                     &                      & 0.230                     & 0.376                     & 0.335                     & 0.203                     \\
                                           &                                                                                       & Pubmed                                                                      & 0.183                     & 0.444                     & 0.298                     & 0.163                     &                      & 0.196                     & 0.438                     & 0.307                     & 0.176                     \\
                                           &                                                                                       & Google                                                                      & 0.375                     & 0.351                     & 0.354                     & 0.288                     &                      & 0.383                     & 0.384                     & 0.387                     & 0.311                     \\ \cmidrule(l){2-12} 
    & \multirow{3}{*}{\rotatebox{90}{\begin{tabular}{c}Llama\\ 405B\end{tabular}}}
        & Wikipedia                                                                   & 0.379                     & 0.367                     & 0.372                     & 0.278                     &                      & 0.340                     & 0.376                     & 0.360                     & 0.260                     \\
                                           &                                                                                       & Pubmed                                                                      & 0.267                     & 0.394                     & 0.330                     & 0.221                     &                      & 0.273                     & 0.404                     & 0.329                     & 0.229                     \\
                                           &                                                                                       & Google                                                                      & 0.434                     & 0.383                     & 0.395                     & 0.321                     &                      & 0.444                     & 0.363                     & 0.396                     & 0.317                     \\ \cmidrule(l){2-12} 
                                           & \multirow{3}{*}{\rotatebox{90}{Phi-4}}                                                & Wikipedia                                                                   & 0.424                     & 0.416                     & 0.402                     & 0.306                     &                      & 0.463                     & 0.438                     & 0.408                     & 0.325                     \\
                                           &                                                                                       & Pubmed                                                                      & 0.308                     & 0.500                     & 0.333                     & 0.230                     &                      & 0.326                     & 0.553                     & 0.330                     & 0.242                     \\
                                           &                                                                                       & Google                                                                      & 0.453                     & 0.385                     & 0.376                     & 0.334                     &                      & 0.495                     & 0.389                     & 0.402                     & 0.360                     \\ \cmidrule(l){2-12} 
                                           & \multirow{3}{*}{\rotatebox{90}{Qwen 2.5}}                                             & Wikipedia                                                                   & 0.238                     & 0.552                     & 0.347                     & 0.228                     &                      & 0.214                     & 0.433                     & 0.321                     & 0.193                     \\
                                           &                                                                                       & Pubmed                                                                      & 0.153                     & 0.404                     & 0.283                     & 0.119                     &                      & 0.173                     & 0.559                     & 0.295                     & 0.141                     \\
                                           &                                                                                       & Google                                                                      & 0.358                     & 0.393                     & 0.383                     & 0.301                     &                      & 0.389                     & 0.407                     & 0.413                     & 0.329                     \\ \cmidrule(l){2-12} 
                                           & \multirow{3}{*}{\rotatebox{90}{Mistral}}                                              & Wikipedia                                                                   & \multicolumn{1}{l}{0.320} & \multicolumn{1}{l}{0.530} & \multicolumn{1}{l}{0.379} & \multicolumn{1}{l}{0.281} & \multicolumn{1}{l}{} & \multicolumn{1}{l}{0.322} & \multicolumn{1}{l}{0.516} & \multicolumn{1}{l}{0.382} & \multicolumn{1}{l}{0.283} \\
                                           &                                                                                       & Pubmed                                                                      & \multicolumn{1}{l}{0.196} & \multicolumn{1}{l}{0.476} & \multicolumn{1}{l}{0.298} & \multicolumn{1}{l}{0.154} & \multicolumn{1}{l}{} & \multicolumn{1}{l}{0.202} & \multicolumn{1}{l}{0.474} & \multicolumn{1}{l}{0.295} & \multicolumn{1}{l}{0.160} \\
                                           &                                                                                       & Google                                                                      & \multicolumn{1}{l}{0.399} & \multicolumn{1}{l}{0.394} & \multicolumn{1}{l}{0.374} & \multicolumn{1}{l}{0.331} & \multicolumn{1}{l}{} & \multicolumn{1}{l}{0.440} & \multicolumn{1}{l}{0.409} & \multicolumn{1}{l}{0.401} & \multicolumn{1}{l}{0.359} \\ 
                                           
                                            \midrule
\multirow{15}{*}{\rotatebox{90}{Liar}} 
    & \multirow{3}{*}{\rotatebox{90}{\begin{tabular}{c}Llama\\ 70B\end{tabular}}}
& Wikipedia                                                                   & 0.210   & 0.234  & 0.185  & 0.114  &  & 0.219           & 0.573           & 0.196             & 0.130           \\
                                           &                                                                                       & Pubmed                                                                      & 0.207   & 0.353  & 0.183  & 0.100  &  & 0.200           & 0.319           & 0.175             & 0.087           \\
                                  &          & Google                                                                      & 0.394   & 0.657  & 0.388  & 0.382  &  & 0.402           & 0.659           & 0.397             & 0.393           \\ \cmidrule(l){2-12} 
                                           
    & \multirow{3}{*}{\rotatebox{90}{\begin{tabular}{c}Llama\\ 405B\end{tabular}}}
        & Wikipedia                                                                   & 0.243   & 0.241  & 0.235  & 0.207  &  & 0.256           & 0.245           & 0.247             & 0.217           \\
                                           &                                                                                       & Pubmed                                                                      & 0.222   & 0.212  & 0.219  & 0.170  &  & 0.203           & 0.188           & 0.198             & 0.143           \\
                                           &                                                                                       & Google                                                                      & 0.396   & 0.514  & 0.398  & 0.401  &  & 0.415           & 0.541           & 0.413             & 0.419           \\ \cmidrule(l){2-12} 
                                           & \multirow{3}{*}{\rotatebox{90}{Phi-4}}                                                & Wikipedia                                                                   & 0.226   & 0.237  & 0.212  & 0.164  &  & 0.230           & 0.258           & 0.212             & 0.167           \\
                                           &                                                                                       & Pubmed                                                                      & 0.202   & 0.208  & 0.183  & 0.108  &  & 0.200           & 0.193           & 0.183             & 0.106           \\
                                           &                                                                                       & Google                                                                      & 0.387   & 0.456  & 0.391  & 0.386  &  & 0.385           & 0.454           & 0.390             & 0.387           \\ \cmidrule(l){2-12} 
                                           & \multirow{3}{*}{\rotatebox{90}{Qwen 2.5}}                                             & Wikipedia                                                                   & 0.210   & 0.334  & 0.184  & 0.096  &  & 0.211           & 0.323           & 0.185             & 0.097           \\
                                           &                                                                                       & Pubmed                                                                      & 0.201   & 0.090  & 0.174  & 0.073  &  & 0.200           & 0.085           & 0.172             & 0.071           \\
                                           &                                                                                       & Google                                                                      & 0.402   & 0.737  & 0.391  & 0.382  &  & 0.408           & 0.728           & 0.397             & 0.389          
                                               \\ \bottomrule
\end{tabular}
\caption{Original-only vs.\ original+negated evidence in zero-shot evaluation. We report Accuracy (A), Precision (P), Recall (R), and macro-$F_{1}$ across datasets, models, and knowledge sources (Wikipedia, PubMed, Google).}

\label{tab:originalvsnegated}
\end{table*}

\begin{table*}[]
\centering
\begin{tabular}{@{}cccccccccccc@{}}
\toprule
\multirow{2}{*}{Dataset}                 & \multirow{2}{*}{Model}                                                                & \multirow{2}{*}{\begin{tabular}[c]{@{}c@{}}Knowledge\\ Source\end{tabular}} & \multicolumn{4}{c}{Original Claim} &  & \multicolumn{4}{c}{Original + Negated Claim}                            \\ \cmidrule(lr){4-7} \cmidrule(l){9-12}

                                         &                                                                                       &                                                                             & A       & P      & R      & F1     &  & A               & P               & R                 & F1              \\ \midrule

\multirow{3}{*}{\rotatebox{90}{Liar}}
      & \multirow{3}{*}{\rotatebox{90}{Mistral}}                                              & Wikipedia                                                                   & 0.210   & 0.334  & 0.184  & 0.096  &  & 0.211           & 0.323           & 0.185             & 0.097           \\
                                           &                                                                                       & Pubmed                                                                      & 0.214   & 0.134  & 0.184  & 0.105  &  & 0.206           & 0.116           & 0.176             & 0.095           \\
                                           &                                                                                       & Google                                                                      & 0.375   & 0.608  & 0.361  & 0.356  &  & 0.381           & 0.631           & 0.365             & 0.362      \\  \midrule

                                         \multirow{15}{*}{\rotatebox{90}{PubHealth}}
    & \multirow{3}{*}{\rotatebox{90}{\begin{tabular}{c}Llama\\ 70B\end{tabular}}}
         & Wikipedia                                                                   & \multicolumn{1}{l}{0.245} & \multicolumn{1}{l}{0.333} & \multicolumn{1}{l}{0.313} & \multicolumn{1}{l}{0.197} & \multicolumn{1}{l}{} & \multicolumn{1}{l}{0.241} & \multicolumn{1}{l}{0.325} & \multicolumn{1}{l}{0.318} & \multicolumn{1}{l}{0.195} \\
                                            &                                                                                       & Pubmed                                                                      & \multicolumn{1}{l}{0.199} & \multicolumn{1}{l}{0.313} & \multicolumn{1}{l}{0.315} & \multicolumn{1}{l}{0.163} & \multicolumn{1}{l}{} & \multicolumn{1}{l}{0.207} & \multicolumn{1}{l}{0.341} & \multicolumn{1}{l}{0.331} & \multicolumn{1}{l}{0.169} \\
                                            &                                                                                       & Google                                                                      & \multicolumn{1}{l}{0.451} & \multicolumn{1}{l}{0.377} & \multicolumn{1}{l}{0.356} & \multicolumn{1}{l}{0.266} & \multicolumn{1}{l}{} & \multicolumn{1}{l}{0.467} & \multicolumn{1}{l}{0.391} & \multicolumn{1}{l}{0.393} & \multicolumn{1}{l}{0.292} \\ \cmidrule(l){2-12} 

    & \multirow{3}{*}{\rotatebox{90}{\begin{tabular}{c}Llama\\ 405B\end{tabular}}}
        & Wikipedia                                                                   & \multicolumn{1}{l}{0.317} & \multicolumn{1}{l}{0.319} & \multicolumn{1}{l}{0.346} & \multicolumn{1}{l}{0.240} & \multicolumn{1}{l}{} & \multicolumn{1}{l}{0.353} & \multicolumn{1}{l}{0.480} & \multicolumn{1}{l}{0.386} & \multicolumn{1}{l}{0.270} \\
                                            &                                                                                       & Pubmed                                                                      & \multicolumn{1}{l}{0.323} & \multicolumn{1}{l}{0.323} & \multicolumn{1}{l}{0.380} & \multicolumn{1}{l}{0.247} & \multicolumn{1}{l}{} & \multicolumn{1}{l}{0.324} & \multicolumn{1}{l}{0.329} & \multicolumn{1}{l}{0.354} & \multicolumn{1}{l}{0.243} \\
                                            &                                                                                       & Google                                                                      & \multicolumn{1}{l}{0.535} & \multicolumn{1}{l}{0.631} & \multicolumn{1}{l}{0.384} & \multicolumn{1}{l}{0.316} & \multicolumn{1}{l}{} & \multicolumn{1}{l}{0.559} & \multicolumn{1}{l}{0.392} & \multicolumn{1}{l}{0.398} & \multicolumn{1}{l}{0.329} \\ \cmidrule(l){2-12} 
                                            & \multirow{3}{*}{\rotatebox{90}{Phi-4}}                                                & Wikipedia                                                                   & \multicolumn{1}{l}{0.305} & \multicolumn{1}{l}{0.364} & \multicolumn{1}{l}{0.314} & \multicolumn{1}{l}{0.255} & \multicolumn{1}{l}{} & \multicolumn{1}{l}{0.330} & \multicolumn{1}{l}{0.383} & \multicolumn{1}{l}{0.329} & \multicolumn{1}{l}{0.275} \\
                                            &                                                                                       & Pubmed                                                                      & \multicolumn{1}{l}{0.261} & \multicolumn{1}{l}{0.307} & \multicolumn{1}{l}{0.283} & \multicolumn{1}{l}{0.213} & \multicolumn{1}{l}{} & \multicolumn{1}{l}{0.284} & \multicolumn{1}{l}{0.328} & \multicolumn{1}{l}{0.288} & \multicolumn{1}{l}{0.230} \\
                                            &                                                                                       & Google                                                                      & \multicolumn{1}{l}{0.492} & \multicolumn{1}{l}{0.421} & \multicolumn{1}{l}{0.355} & \multicolumn{1}{l}{0.336} & \multicolumn{1}{l}{} & \multicolumn{1}{l}{0.517} & \multicolumn{1}{l}{0.413} & \multicolumn{1}{l}{0.365} & \multicolumn{1}{l}{0.356} \\ \cmidrule(l){2-12} 
                                            & \multirow{3}{*}{\rotatebox{90}{Qwen 2.5}}                                             & Wikipedia                                                                   & \multicolumn{1}{l}{0.183} & \multicolumn{1}{l}{0.343} & \multicolumn{1}{l}{0.314} & \multicolumn{1}{l}{0.158} & \multicolumn{1}{l}{} & \multicolumn{1}{l}{0.194} & \multicolumn{1}{l}{0.345} & \multicolumn{1}{l}{0.322} & \multicolumn{1}{l}{0.170} \\
                                            &                                                                                       & Pubmed                                                                      & \multicolumn{1}{l}{0.165} & \multicolumn{1}{l}{0.303} & \multicolumn{1}{l}{0.308} & \multicolumn{1}{l}{0.139} & \multicolumn{1}{l}{} & \multicolumn{1}{l}{0.176} & \multicolumn{1}{l}{0.322} & \multicolumn{1}{l}{0.304} & \multicolumn{1}{l}{0.151} \\
                                            &                                                                                       & Google                                                                      & \multicolumn{1}{l}{0.463} & \multicolumn{1}{l}{0.473} & \multicolumn{1}{l}{0.387} & \multicolumn{1}{l}{0.281} & \multicolumn{1}{l}{} & \multicolumn{1}{l}{0.476} & \multicolumn{1}{l}{0.391} & \multicolumn{1}{l}{0.386} & \multicolumn{1}{l}{0.295} \\ \cmidrule(l){2-12} 
                                            & \multirow{3}{*}{\rotatebox{90}{Mistral}}                                              & Wikipedia                                                                   & \multicolumn{1}{l}{0.424} & \multicolumn{1}{l}{0.297} & \multicolumn{1}{l}{0.266} & \multicolumn{1}{l}{0.229} & \multicolumn{1}{l}{} & \multicolumn{1}{l}{0.417} & \multicolumn{1}{l}{0.274} & \multicolumn{1}{l}{0.247} & \multicolumn{1}{l}{0.212} \\
                                            &                                                                                       & Pubmed                                                                      & \multicolumn{1}{l}{0.425} & \multicolumn{1}{l}{0.262} & \multicolumn{1}{l}{0.223} & \multicolumn{1}{l}{0.184} & \multicolumn{1}{l}{} & \multicolumn{1}{l}{0.440} & \multicolumn{1}{l}{0.302} & \multicolumn{1}{l}{0.239} & \multicolumn{1}{l}{0.198} \\
                                            &                                                                                       & Google                                                                      & \multicolumn{1}{l}{0.496} & \multicolumn{1}{l}{0.349} & \multicolumn{1}{l}{0.304} & \multicolumn{1}{l}{0.278} & \multicolumn{1}{l}{} & \multicolumn{1}{l}{0.488} & \multicolumn{1}{l}{0.341} & \multicolumn{1}{l}{0.283} & \multicolumn{1}{l}{0.258} \\ \bottomrule
\end{tabular}
\caption{Extension of Table \ref{tab:originalvsnegated}: Comparison of Original Claims vs. Original + Negated Claims }
\label{tab:orginalvsnegated2}
\end{table*}

\begin{table*}[]
\centering
\begin{tabular}{@{}ccccccccccc@{}}
\toprule
\multirow{2}{*}{Dataset}   &  & \multirow{2}{*}{Model} &  & \multicolumn{7}{c}{Knowledge Source}                 \\ \cmidrule(l){5-11} 
                           &  &                        &  & Wikipedia &  & Pubmed &  & Google &  & Merged(W+P+G) \\ \cmidrule(r){1-1} \cmidrule(lr){3-3} \cmidrule(lr){5-5} \cmidrule(lr){7-7} \cmidrule(lr){9-9} \cmidrule(l){11-11} 
\multirow{5}{*}{SciFact}   &  & Llama 70B              &  & 0.430     &  & 0.597  &  & 0.550  &  & 0.610         \\
                           &  & Llama 405B             &  & 0.447     &  & 0.593  &  & 0.580  &  & 0.597         \\
                           &  & Phi-4                  &  & 0.410     &  & 0.583  &  & 0.590  &  & 0.583         \\
                           &  & Qwen 2.5               &  & 0.437     &  & 0.583  &  & 0.587  &  & 0.607         \\
                           &  & Mistral                &  & 0.393     &  & 0.573  &  & 0.583  &  & 0.617         \\ \midrule
\multirow{5}{*}{Averitec}  &  & Llama 70B              &  & 0.230     &  & 0.196  &  & 0.383  &  & 0.384         \\
                           &  & Llama 405B             &  & 0.340     &  & 0.273  &  & 0.444  &  & 0.574         \\
                           &  & Phi-4                  &  & 0.463     &  & 0.326  &  & 0.495  &  & 0.515         \\
                           &  & Qwen 2.5               &  & 0.214     &  & 0.173  &  & 0.389  &  & 0.387         \\
                           &  & Mistral                &  & 0.322     &  & 0.202  &  & 0.440  &  & 0.521         \\ \midrule
\multirow{5}{*}{Liar}      &  & Llama 70B              &  & 0.219     &  & 0.200  &  & 0.402  &  & 0.300         \\
                           &  & Llama 405B             &  & 0.256     &  & 0.203  &  & 0.415  &  & 0.320         \\
                           &  & Phi-4                  &  & 0.230     &  & 0.200  &  & 0.385  &  & 0.289         \\
                           &  & Qwen 2.5               &  & 0.211     &  & 0.200  &  & 0.408  &  & 0.285               \\
                           &  & Mistral                &  & 0.211     &  & 0.206  &  & 0.381  &  & 0.294         \\ \midrule
\multirow{5}{*}{PubHealth} &  & Llama 70B              &  & 0.241     &  & 0.207  &  & 0.467  &  & 0.449         \\
                           &  & Llama 405B             &  & 0.353     &  & 0.324  &  & 0.559  &  & 0.557               \\
                           &  & Phi-4                  &  & 0.330     &  & 0.284  &  & 0.517  &  & 0.553              \\
                           &  & Qwen 2.5               &  & 0.194     &  & 0.176  &  & 0.476  &  & 0.467         \\
                           &  & Mistral                &  & 0.417     &  & 0.440  &  & 0.488  &  & 0.490         \\ \bottomrule
\end{tabular}
\caption{Comparative Accuracy of Individual vs. Aggregated Knowledge Sources (Wikipedia+PubMed+Google)}
\label{tab:individualvsaggregated}
\end{table*}

\section{Automated Dual-Perspective Multi-Source Retrieval-Based Claim Verification}

Our novel ODCV approach systematically generates, retrieves, selects, and evaluates evidence to accurately verify claims.  Figure ~\ref{fig:proposed_system} shows an overview of our architecture which consists of three key components: (1) negated claim generation, (2) evidence retrieval and selection, and (3) claim verification. We describe each component in detail below. 

\subsection{Datasets}
We evaluate our ODCV system on four benchmarks ranging scientific, health, sociopolitical, and political claims: \textbf{SciFact}~\cite{wadden2020fact} (biomedical claims with sentence-level evidence from abstracts), \textbf{PubHealth}~\cite{kotonya2020explainable} (health claims with expert justifications), \textbf{Averitec}~\cite{schlichtkrull2023averitec} (controversial/ambiguous sociopolitical claims emphasizing uncertainty), and \textbf{LIAR}~\cite{wang2017liar} (large-scale political claims from speeches, social media, and news). Table~\ref{tab:dataset_description} in the appendix summarizes domains, sources, label sets, and claim counts.

\begin{center}
   \begin{figure*}
        \centering
        \includegraphics[width=\linewidth]{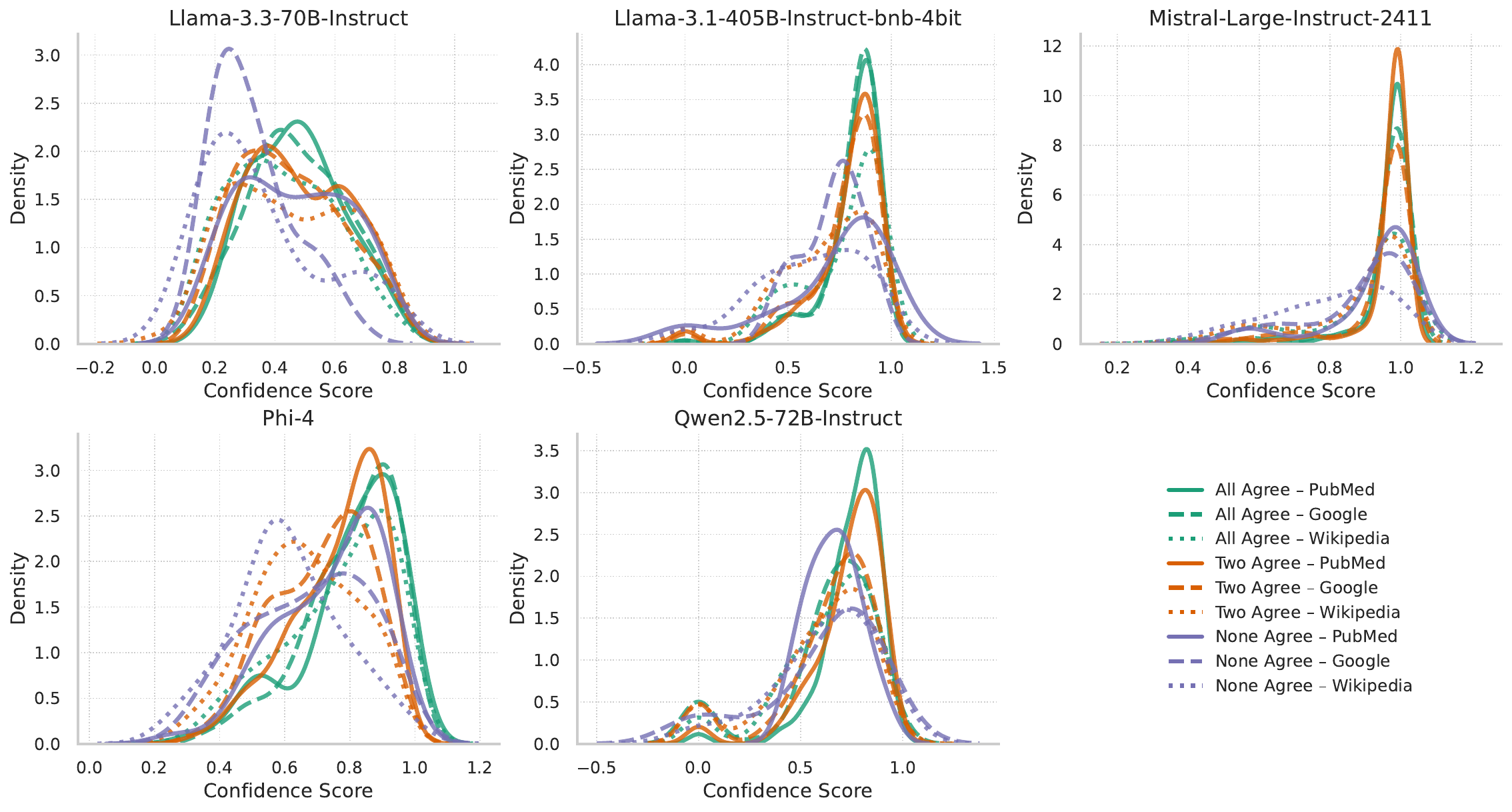} % Adjust the width
        \caption{Confidence distribution (KDE) across  knowledge sources for the Averitec dataset, illustrating variation in model certainty and inter-source disagreement for claim verification}
        \label{fig:Averitec_disagreement}
    \end{figure*}
\end{center}

\subsection{Negated Claim Generation}

For each claim set $\mathcal{C}=\{c_1,\dots,c_n\}$, we generate a negated counterpart $\bar{c}_i$ for every $c_i$ using Mistral AI\footnote{https://docs.mistral.ai/api/}, producing informative contrasts (including numerical reframing). For example: \emph{“A deficiency of vitamin B12 increases homocysteine”} $\rightarrow$ \emph{“A surplus of vitamin B12 decreases homocysteine”}; and \emph{“5\% of perinatal mortality is due to low birth weight”} $\rightarrow$ \emph{“95\% of perinatal mortality is not due to low birth weight”}. Pairing $(c_i,\bar{c}_i)$ ensures both supportive and contradictory perspectives for subsequent retrieval and verification.

\subsection{Evidence Retrieval}

We use three major knowledge sources, Wikipedia, PubMed, and Google, guided by \cite{vladika2024comparing}, which finds Wikipedia stronger on popular/trending claims and PubMed more precise for technical/scientific queries. To ensure coverage and domain adaptability, we utilize all three.

Let $\mathcal{K}=\{k_1,k_2,k_3\}$ denote Wikipedia, PubMed, and Google. For each claim $c_i\in\mathcal{C}$ and source $k_j\in\mathcal{K}$, we retrieve the top-$k$ documents
\[
R(c_i,k_j)=\{d_1^{(i,j)},\dots,d_k^{(i,j)}\},
\]
and do so for both $c_i$ and its negation $\bar{c}_i$ to gather supportive, neutral, and potentially contradictory evidence.

\textbf{Source-specific pipelines:}
\begin{itemize}
\item \textbf{Wikipedia:} We used English dumps\footnote{https://dumps.wikimedia.org/} ($\sim$7M articles) cleaned with WikiExtractor~\cite{Wikiextractor2015} and indexed in Elasticsearch\footnote{https://www.elastic.co/elasticsearch} for scalable retrieval.
\item \textbf{PubMed:} We used 23.6M abstracts\footnote{https://pubmed.ncbi.nlm.nih.gov/download/} preprocessed and encoded with transformer-based sentence embeddings for dense retrieval, enhanced by BM25~\cite{amati2009bm25} for lexical ranking.
\item \textbf{Google:} We leveraged Google Custom Search API\footnote{https://developers.google.com/custom-search/v1/overview} which queries each claim and returns ranked results (title, snippet, URL) as web evidence.
\end{itemize}

\subsection{Evidence Selection}
% Sentence filtering with SPICED (single-equation version)
After retrieval, we filter sentences using SPICED embeddings~\cite{shushkevich2023spiced}. For each claim \(x \in \{c,\bar{c}\}\), we embed the claim and every sentence from the top \(M\) retrieved documents and compute their cosine similarity. From each document we keep the top-\(k\) sentence(s) by similarity and unite them across documents to form the evidence set. We apply the same procedure to \(c\) and \(\bar{c}\) to capture both supportive and contradictory evidence.

\subsection{Evidence Deduplication and Final Selection}

The core idea is to remove overlapping (duplicate) evidence sentences and merge the rest. For claim $c_i$ and source $k_j$, let $E^+_{ij}=E(c_i,k_j)$ and $E^-_{ij}=E(\bar c_i,k_j)$. After normalization (lowercasing, punctuation stripping), we form candidates via symmetric difference with light merging (to fuse split segments, e.g., \texttt{[SEP]}):
\[
E^{\mathrm{cand}}_{ij}=\mathrm{Merge}\!\big(\tilde E^+_{ij}\triangle \tilde E^-_{ij}\big).
\]
We then rank $E^{\mathrm{cand}}_{ij}$ by SPICED~\cite{shushkevich2023spiced} similarity to $c_i$ and keep the top-$p$ as $E^{\mathrm{final}}_{ij}$. Finally, we aggregate per-claim evidence across sources as
\[
E_i \;=\; \bigcup_{j=1}^{|\mathcal{K}|} E^{\mathrm{final}}_{ij}.
\]

\subsection{Veracity Prediction}

For final claim verification, we employ a large language model (LLM), denoted $L$, to predict the veracity of each claim $c_i$ given $E_i$. We evaluate five state-of-the-art LLMs (open or widely available): Llama~3.3 (70B) and Llama~3.1 (405B)~\cite{grattafiori2024llama}, Mistral-Large, Qwen~2.5~\cite{team2024qwen2}, and Phi-4~\cite{abdin2024phi}, selected for strong benchmark performance and accessibility~\cite{open-llm-leaderboard-v2}. In a zero-shot setting, each model is prompted with the claim and its evidence using a direct claim verification prompt ( Appendix Figure~\ref{fig:Prompt}) adapted to each model and dataset.

We present veracity as $m$-class prediction over $Y=\{y_1,\dots,y_m\}$ (e.g., $m=3$: \textit{Refuted}, \textit{Supported}, \textit{Not Enough Evidence}). For each claim--evidence pair, the LLM outputs
\[
\hat y_i \;=\; F(c_i, E_i; L) \in Y,
\]
via a single-token choice (e.g., A–C) deterministically mapped to dataset classes.

\subsection{Measuring Source‑Level Disagreement}
Recent work shows that token-level log-probabilities (“logprobs”) help interpret LLM decisions~\cite{kauf2024log}. We treat each veracity label as a discrete token and use the logprob of the chosen label as the confidence score. For claim $c_i$ with evidence $E_i$, letting $\mathbf z$ be the logits over labels $\mathcal{Y}$ and $\hat y_i=\arg\max\mathrm{softmax}(\mathbf z)$, confidence is
\[
\mathrm{conf}(c_i,E_i)=\log\!\big(\mathrm{softmax}(\mathbf z)\big)_{\hat y_i}.
\]
We compute this per source and compare the logprobs across sources: low dispersion indicates agreement, while high dispersion signals disagreement and uncertainty; visualizations make these patterns explicit.

\section{Results}

We evaluate in a zero-shot setting (no fine-tuning) to isolate the contribution of dual-perspective multi-source method rather than optimize absolute scores. Across SciFact, Averitec, LIAR, and PubHealth datasets with five LLMs (Llama 70B, Llama 405B, Phi-4, Qwen 2.5, Mistral), two robust trends emerge (Tables~\ref{tab:originalvsnegated}, \ref{tab:orginalvsnegated2}, \ref{tab:individualvsaggregated}). 

\emph{First}, augmenting each claim with its explicit negation generally improves accuracy and macro-$F_1$ over using the original claim alone, with typical relative gains of about +2--10\% (accuracy) and +2--8\% ($F_1$). Representative examples include SciFact with Llama~70B{+}Google (+10.4\% accuracy, +8.1\% $F_1$; 0.550$\!\rightarrow\!$0.607, 0.530$\!\rightarrow\!$0.573), Averitec with Phi-4{+}Wikipedia (+9.2\%, +6.2\%), LIAR with Llama~405B{+}Google (+4.8\%, +4.5\%), and PubHealth with Phi-4{+}Google (+5.1\%, +6.0\%). While a few source--LLM pairs show neutral or slight decreases (e.g., SciFact with Phi-4{+}Google $F_1$), the overall effect is consistently positive.

\emph{Second}, aggregating evidence from Wikipedia, PubMed, and Google typically boosts performance beyond any single source, especially relative to weaker sources: on SciFact, Llama~70B’s merged $F_1$ exceeds Wikipedia by +41.9\% and Google by +10.9\%; on Averitec, Llama~405B’s merged $F_1$ is +68.8\% over Wikipedia and +29.3\% over Google; on PubHealth, Phi-4’s merged $F_1$ is +67.6\% over Wikipedia and +7.0\% over Google. For LIAR, where Google alone is already strong, merged performance remains above Wikipedia and PubMed but is below Google (e.g., Llama~405B: 0.320 vs.\ 0.256/0.203/0.415). Collectively, dual-perspective retrieval and multi-source aggregation provide complementary, often double-digit relative gains across LLMs and knowledge sources in zero-shot conditions, demonstrating the robustness and practical effectiveness of the proposed approach.

\subsection{Consensus and Conflict Across Sources}

We visualize per‑source confidence (token log‑probability of the predicted veracity label with KDEs formed by source agreement among PubMed, Google, and Wikipedia (\emph{all}, \emph{two}, \emph{none}). \textbf{Averitec} (Figure ~\ref{fig:Averitec_disagreement}) shows the expected ordering: unanimity yields sharper, higher‑confidence peaks; partial agreement is broader and lower; and no agreement is lowest and most dispersed, though curves are flatter than in structured domains. Appendix figures for \textbf{LIAR}, \textbf{PubHealth}, and \textbf{SciFact}  (Figure \ref{fig:Liar_disagreement}, \ref{fig:Pubhealth_disagreement}, \ref{fig:Scifact_disagreement}) confirm the same trend, with clearer separation in SciFact/PubHealth and greater dissemination in LIAR. Confidence magnitudes are sometimes LLM‑dependent: \textbf{Llama} separates agreement regimes most distinctly, \textbf{Mistral} is similar but broader, \textbf{Phi‑4} spreads more under disagreement, and \textbf{Qwen 2.5} shows tight peaks under unanimity. PubHealth violin plots (Appendix Figure ~\ref{fig:Pubhealth_disagreement_distribution}) support inter‑LLM shifts in central tendency and dispersion. Reporting per‑source log‑probabilities and their dispersion alongside the predicted veracity thus quantifies source‑level disagreement (e.g., Google agrees while Wikipedia does not) and makes residual uncertainty transparent.

\section{Discussion}

Dual‑perspective retrieval, considering both the original claim and its negation together with aggregation across sources, yields consistent gains in our zero‑shot results. Kernel Density Estimations (KDEs) of per-source log-probabilities indicate a strong correlation between agreement and certainty: complete agreement across sources like PubMed, Google, and Wikipedia results in sharp peaks of high confidence, whereas partial or no consensus produces lower and wider distributions, more so in open-domain data sets (e.g., LIAR, Averitec) than in structured ones (e.g., SciFact, PubHealth). Explicitly negating claims systematically enhances claim verification by retrieving evidence that both supports and refutes, and combining information from multiple sources further improves performance while decreasing uncertainty when sources are in agreement. Since no single source predominates, the utility of a source is dependent on the claim and domain, justifying aggregation. In practice, displaying per-source confidence and its distribution offers a model-agnostic indicator of reliability and reveals disagreements transparently for end users. However, because raw confidence levels vary between LLMs, comparisons across them require calibration or intra-model baselines. Remaining challenges include time-sensitive evidence and handling long contexts, which may hinder certainty even when aggregation is employed.

\section{Future Work}

Validated in zero-shot settings, our next steps are to combine dual-perspective multi-source claim verification with more sophisticated approaches. We will add temporal reasoning with time‑aware retrieval and alignment of claims and evidence. We will extend to multilingual verification and aim for comparable performance across languages. We will develop context‑aware retrieval that adapts to user context under neutrality constraints. We will improve web evidence quality by grading content and estimating source reliability to prioritize stronger evidence. We will mitigate hallucinations through faithfulness checks, confidence calibration, and enforcing consistency between evidence and veracity predictions.

\section{Conclusions}

We presented an open-domain claim verification system that retrieves with both original and negated claims to capture support and contradiction, aggregates evidence from Wikipedia, PubMed, and Google, and quantifies uncertainty via label log-probabilities and KDE-based visualizations. Through a series of experiments, we showed that negated-claim retrieval and multi-source aggregation yield consistent, complementary gains without fine-tuning, improving performance. Visualizations of log-probabilities improve  interpretability of veracity predictions. 

Llama and Mistral showed consistently strong performance across knowledge sources and datasets. By exposing source-level agreement and confidence score, we strengthen interpretability of veracity predictions.

\section*{Limitations}
Our study has several limitations. First, constrained context windows can truncate or underweight relevant passages when verification requires long, multi‑document evidence; although emerging long‑context models (e.g., $\geq$32K tokens) may help in this, we did not exploit them here, and hierarchical selection/ordering effects (“lost in the middle”) may depress performance. Second, open‑domain datasets such as \emph{LIAR} include noisy or adversarial claims and minority labels; in zero‑shot classification, this combination yields lower macro‑$F_{1}$ even when accuracy is moderate, and models struggle to resolve incomplete or genuinely conflicting evidence. Third, we do not impose time‑aware retrieval or reasoning, so evidence that is outdated or post‑dates the claim can lead to inconsistent veracity predictions for time‑sensitive statements. Finally, like other LLM-based systems, our approach remains sensitive to distribution shifts, misleading inputs, and biases in source corpora and pre-training, which can limit generalization and reliability in real-world settings.

% \section*{Acknowledgments}

% Bibliography entries for the entire Anthology, followed by custom entries
% \bibliography{anthology,custom}
% Custom bibliography entries only
\bibliography{custom}
\newpage
\appendix
\section{Appendix}
\subsection{Dataset Description (Table \ref{tab:dataset_description})}

\label{sec:appendix}

\begin{table*}[]
\begin{tabular}{@{}ccccc@{}}
\toprule
\textbf{Dataset}           & \textbf{Source}                 & \textbf{Domain}             & \textbf{Label}                     & \textbf{Claims}         \\ \midrule
\multirow{3}{*}{SCIFACT}   & \multirow{3}{*}{Science}        & \multirow{3}{*}{Scientific} & Supported                          & \multirow{3}{*}{1400}   \\
                           &                                 &                             & Refuted                            &                         \\
                           &                                 &                             & Not Enough Info                    &                         \\ \midrule
\multirow{4}{*}{Averitec}  & \multirow{4}{*}{Factcheck}      & \multirow{4}{*}{Mixed}      & Supported                          & \multirow{4}{*}{4568}   \\
                           &                                 &                             & Refuted                            &                         \\
                           &                                 &                             & Conflicting evidence/cherrypicking &                         \\
                           &                                 &                             & Not Enough Info                    &                         \\ \midrule
\multirow{6}{*}{LIAR}      & \multirow{6}{*}{POLITIFACT.COM} & \multirow{6}{*}{Fake News}  & Pants on Fire                      & \multirow{6}{*}{12,836} \\
                           &                                 &                             & False                              &                         \\
                           &                                 &                             & Barely True                        &                         \\
                           &                                 &                             & Half True                          &                         \\
                           &                                 &                             & Mostly True                        &                         \\
                           &                                 &                             & True                               &                         \\ \midrule
\multirow{4}{*}{PUBHEALTH} & \multirow{4}{*}{Factcheck}      & \multirow{4}{*}{Biomedical} & True                               & \multirow{4}{*}{11,832} \\
                           &                                 &                             & False                              &                         \\
                           &                                 &                             & Mixture                            &                         \\
                           &                                 &                             & Unproven                           &                         \\ \bottomrule
\end{tabular}
\caption{Overview of Benchmark Datasets}
\label{tab:dataset_description}
\end{table*}

\begin{center}
   \begin{figure*}
        \centering
        \includegraphics[width=\linewidth]{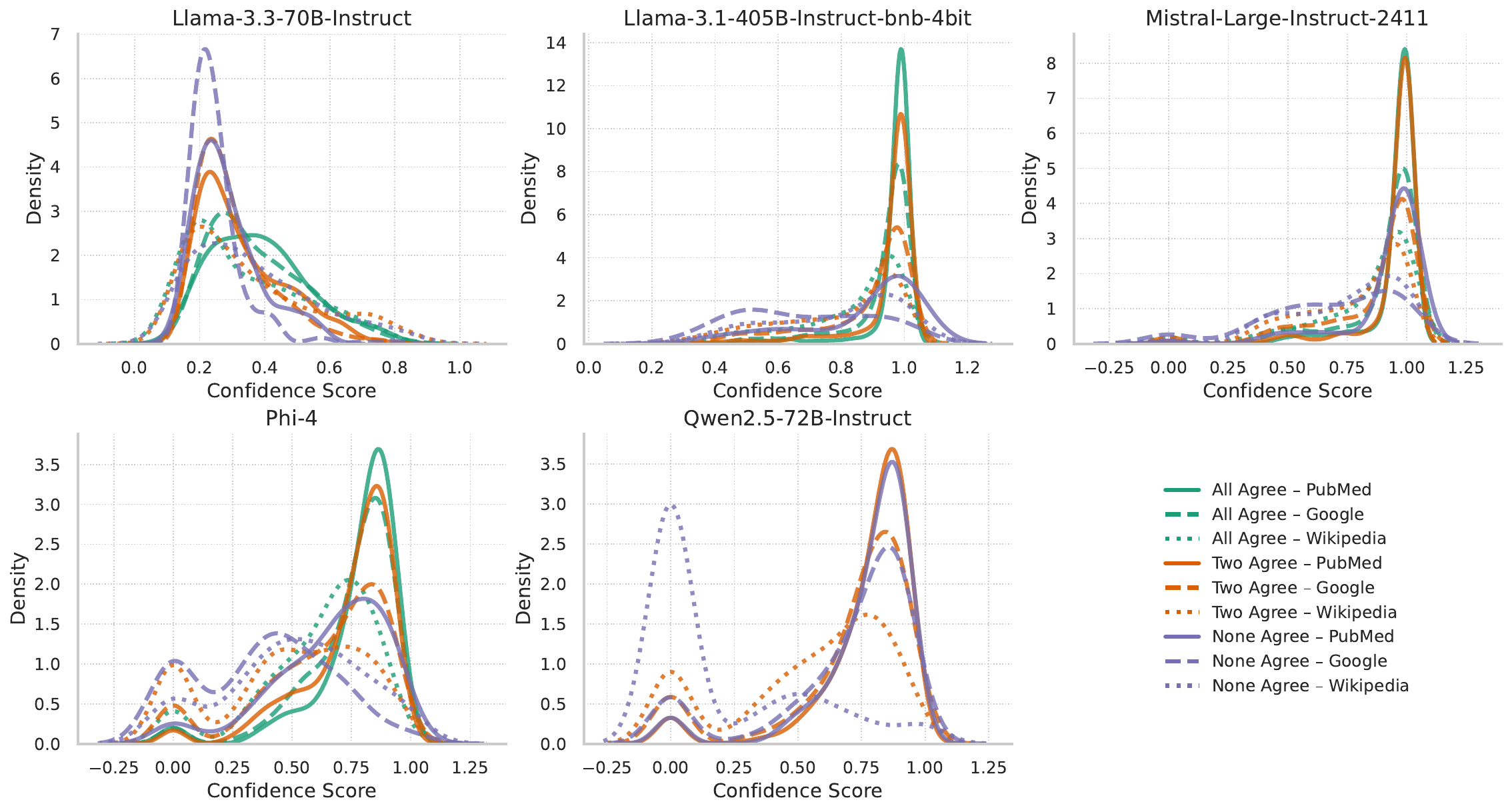} % Adjust the width
        \caption{Confidence distribution (KDE) across different knowledge sources for the LIAR dataset, illustrating variation in model certainty and inter-source disagreement during claim verification}
        \label{fig:Liar_disagreement}
    \end{figure*}
\end{center}

\subsection{Disagreement for LIAR (Figure \ref{fig:Liar_disagreement})}

\begin{center}
   \begin{figure*}
        \centering
        \includegraphics[width=\linewidth]{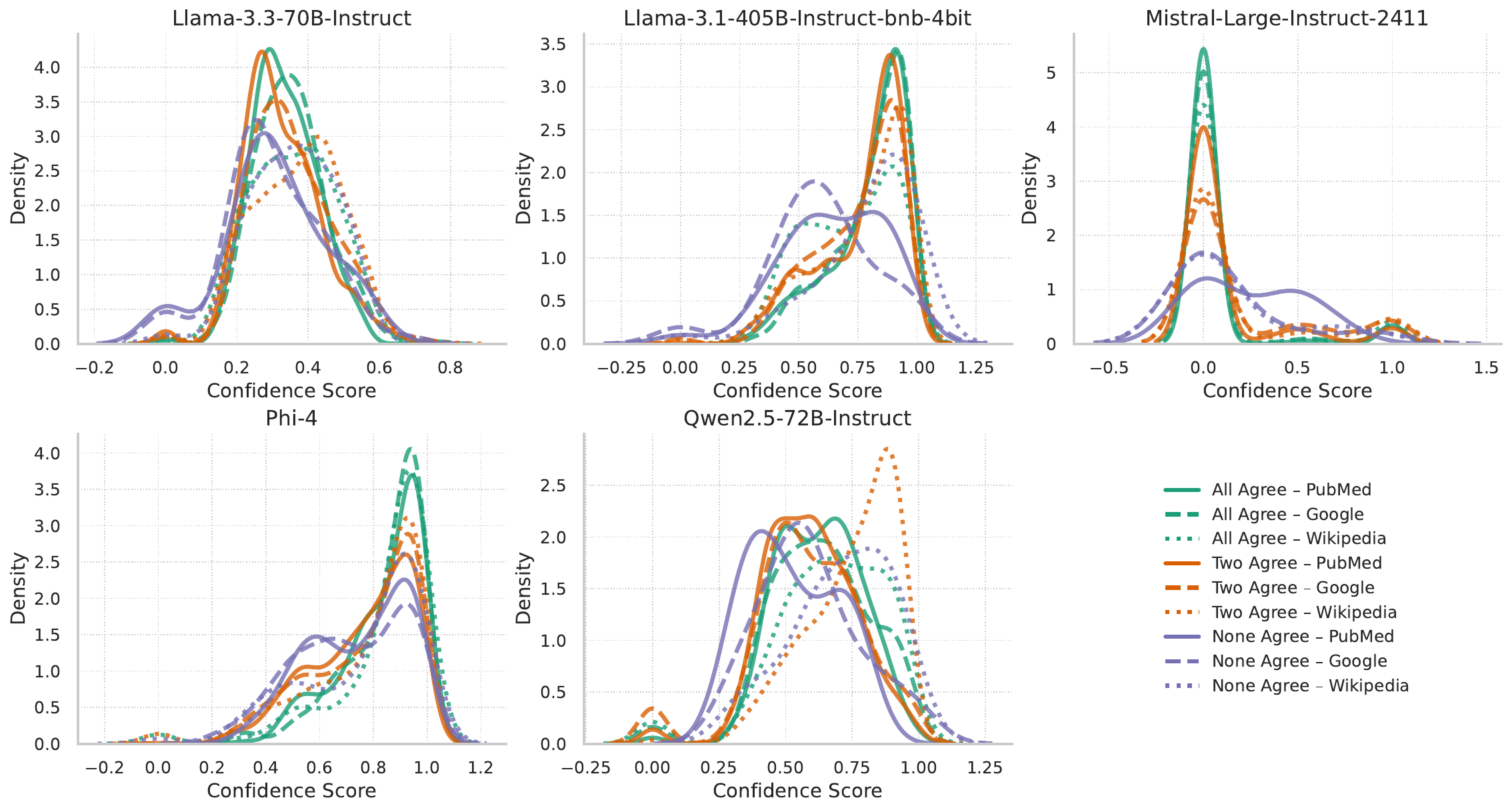} % Adjust the width
        \caption{Confidence distribution (KDE) across different knowledge sources for the Pubhealth dataset, illustrating variation in model certainty and inter-source disagreement during claim verification}
        \label{fig:Pubhealth_disagreement}
    \end{figure*}
\end{center}

\subsection{Disagreement for Pubhealth (Figure \ref{fig:Pubhealth_disagreement})}

\begin{center}
   \begin{figure*}
        \centering
        \includegraphics[width=\linewidth]{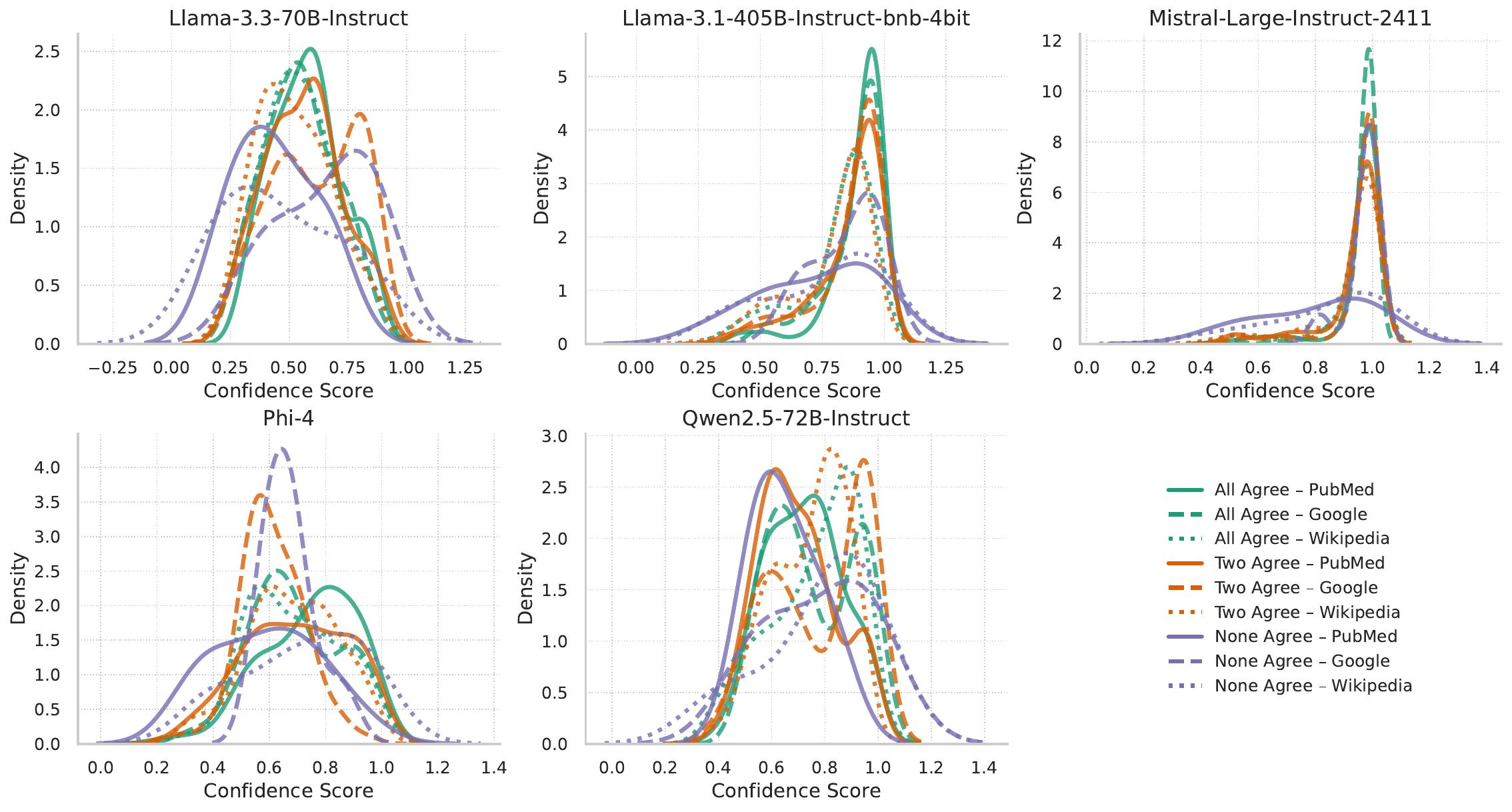} % Adjust the width
        \caption{Confidence distribution (KDE) across different knowledge sources for the SCIFact dataset, illustrating variation in model certainty and inter-source disagreement during claim verification}
        \label{fig:Scifact_disagreement}
    \end{figure*}
\end{center}

\subsection{Disagreement for Pubhealth (Figure \ref{fig:Scifact_disagreement})}

\begin{center}
   \begin{figure}
        \centering
        \includegraphics[width=\linewidth]{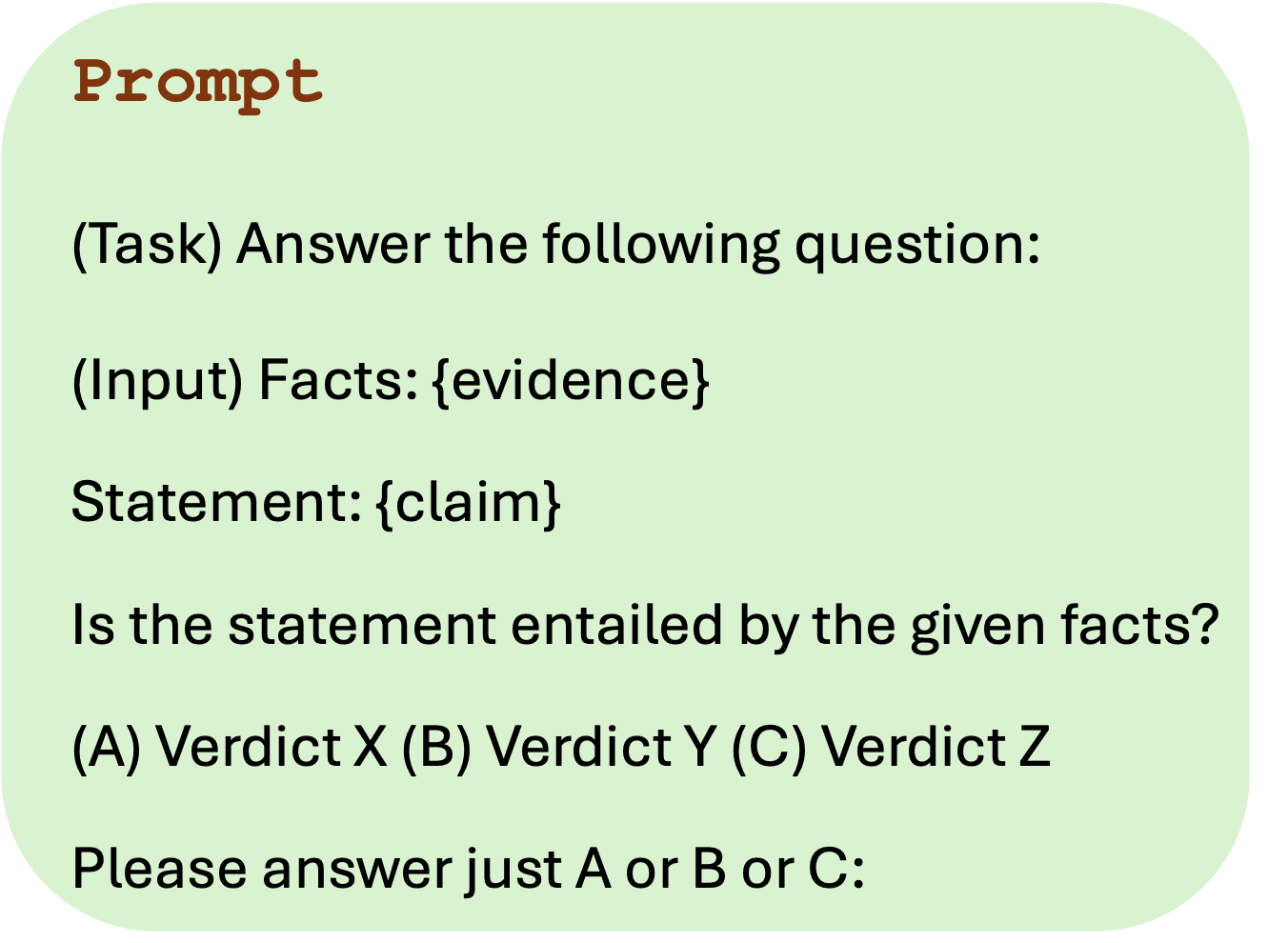} % Adjust the width
        \caption{Prompt template for veracity prediction}
        \label{fig:Prompt}
    \end{figure}
\end{center}

\subsection{Prompt for veracity prediction (Figure \ref{fig:Prompt})}

\begin{center}
   \begin{figure}
        \centering
        \includegraphics[width=\linewidth]{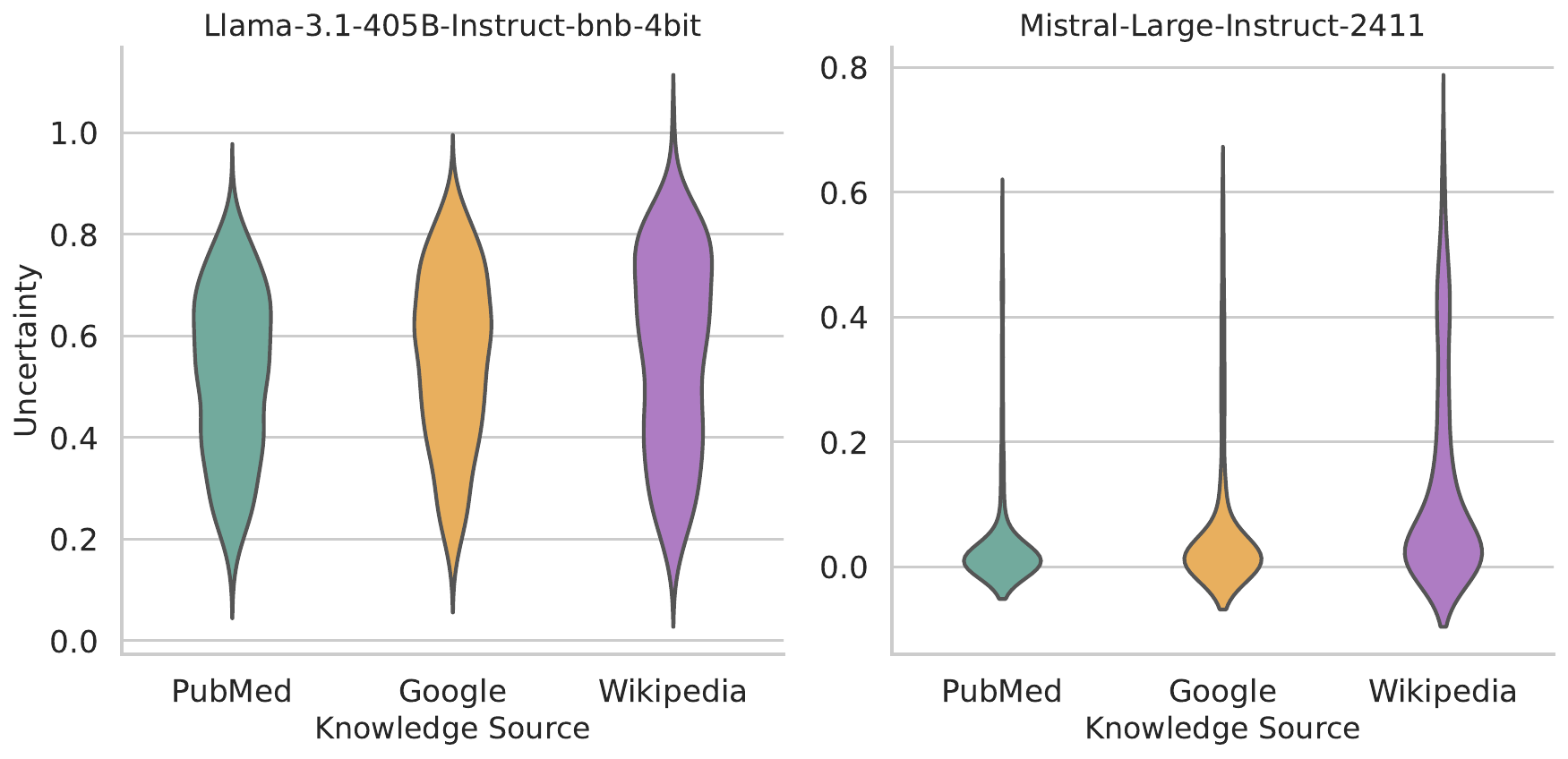} % Adjust the width
        \caption{Violin Plot for Pubhealth}
        \label{fig:Pubhealth_disagreement_distribution}
    \end{figure}
\end{center}

\subsection{Pubhealth Confidence score distribution (Figure \ref{fig:Pubhealth_disagreement_distribution})}

\end{document}